\documentclass{article} 
\usepackage{iclr2023_conference,times}
\usepackage{booktabs}
\usepackage{multirow}
\usepackage{rotating}

\usepackage{amsmath,amsfonts,bm}

\def\eqref#1{equation~\ref{#1}}

\def\1{\bm{1}}

\DeclareMathAlphabet{\mathsfit}{\encodingdefault}{\sfdefault}{m}{sl}
\SetMathAlphabet{\mathsfit}{bold}{\encodingdefault}{\sfdefault}{bx}{n}

\usepackage{hyperref}
\usepackage{url}
\usepackage{graphicx}
\usepackage{lipsum}

\graphicspath{ {./images/} }

\title{In-situ Anomaly Detection in Additive \\ Manufacturing with Graph Neural Networks}

\author{Sebastian Larsen \& Paul A. Hooper \\ 
Department of Mechanical Engineering\\
Imperial College London\\
London, SW7 2BX, UK \\
\texttt{\{sebastian.larsen16, paul.hooper\}@imperial.ac.uk}
}

\iclrfinalcopy 
\begin{document}

\maketitle

\begin{abstract}

Transforming a design into a high-quality product is a challenge in metal additive manufacturing due to rare events which can cause defects to form. Detecting these events in-situ could, however, reduce inspection costs, enable corrective action, and is the first step towards a future of tailored material properties. In this study a model is trained on laser input information to predict nominal laser melting conditions. An anomaly score is then calculated by taking the difference between the predictions and new observations. The model is evaluated on a dataset with known defects achieving an F1 score of 0.821. This study shows that anomaly detection methods are an important tool in developing robust defect detection methods.

\end{abstract}

\section{Introduction}
Machine learning (ML) methods are increasingly used in additive manufacturing (AM) to automate inspection \citep{scime2018multi}, predict defects \citep{larsen2022deep}, as well as potentially tailor microstructure and material properties \citep{debroy2021metallurgy}. In Laser-Powder Bed Fusion (L-PBF), a thin layer of powdered material is spread across a build plate. A laser then scans back and forth melting the powder. The process is repeated layer-by-layer until a 3-dimensional part is formed. As a result of chaotic melting conditions, micron sized flaws can occur. Hence, in-situ monitoring cameras have been combined with ML to observe the process. There is, however, a high cost associated with collecting labelled data for training ML algorithms. This requires X-ray computed tomography (XCT) to characterise the flaws which form as pores inside the material. XCT is also limited to very small parts. It is, therefore, desirable to develop robust anomaly detection approaches which require minimal flaws for training ML models.

Another challenge is to represent the AM process complexity with the available information. This can be addressed by geometric deep learning methods. Graph neural networks (GNN) have become increasingly popular for modelling data that takes on a graph type structure. This includes molecules \citep{gainza2020deciphering}, physics simulations \citep{sanchez2020learning}, traffic prediction \citep{derrow2021eta}, and material property prediction \citep{xie2018crystal}. Monitoring signals are dependent on the geometry of the component being manufactured as well as the laser scanning strategy. Therefore, a graph structure is highly suited to modelling the underlying process physics.

The objective of this study is to develop an anomaly detection approach that requires no labelled data other than nominal processing conditions. Due to spatio-temporal correlations in the data, a graph structure is selected to represent the positional information. Given a graph and the laser feature inputs, such as power and scan direction, the model is trained to predict nominal melting conditions. This serves as an approach to compare new observations to the expected result, yielding an anomaly score.

\section{Methodology}

\subsection{Sensor and Data collection}

\begin{figure}[h]
\begin{center}
\includegraphics[width=12cm]{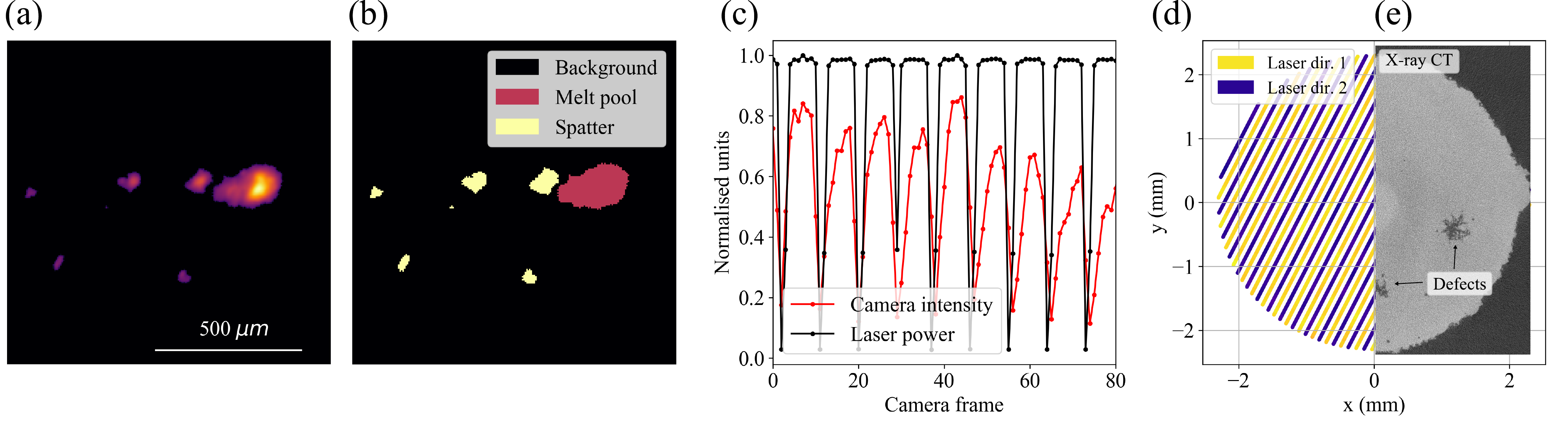}
\end{center}
\caption{(a) Representative image of the melt pool which forms as the laser melts material. (b) Segmentated image. (c) Temporal signal of a melt pool and laser power. (d) Laser scan direction feature. (e) X-ray CT scan image slice with pore defects shown as dark spots.}
\label{fig:data}
\end{figure}

Images were acquired of the melt pool with a high-speed recording at 100 kHz (Fig. \ref{fig:data}a) \citep{hooper2018melt}. The camera is located co-axial to the laser and measures the melt pool radiation. The images were segmented to extract a number of melt pool features including: size, shape, intensity, and number of spatter particles (Fig. \ref{fig:data}b). These features are used as node labels described in the next section. The melt pool signal correlates with the laser power which is pulse width modulated (Fig. \ref{fig:data}c). The laser also changes direction as it scans over the layer (Fig. \ref{fig:data}d). A number of defects were injected into a part during processing and the measured signal taken from the next layer. After the component was built, an X-ray CT scan was carried out (Fig. \ref{fig:data}e). Defects, given by porosity, were characterised within the part using a segmentation algorithm. A small amount of anomalous process conditions could then be mapped back to the measured signal location to act as an evaluation dataset for the anomaly detector.

\subsection{Machine Learning}

\begin{figure}[httb!]
\begin{center}
\includegraphics[width=12cm]{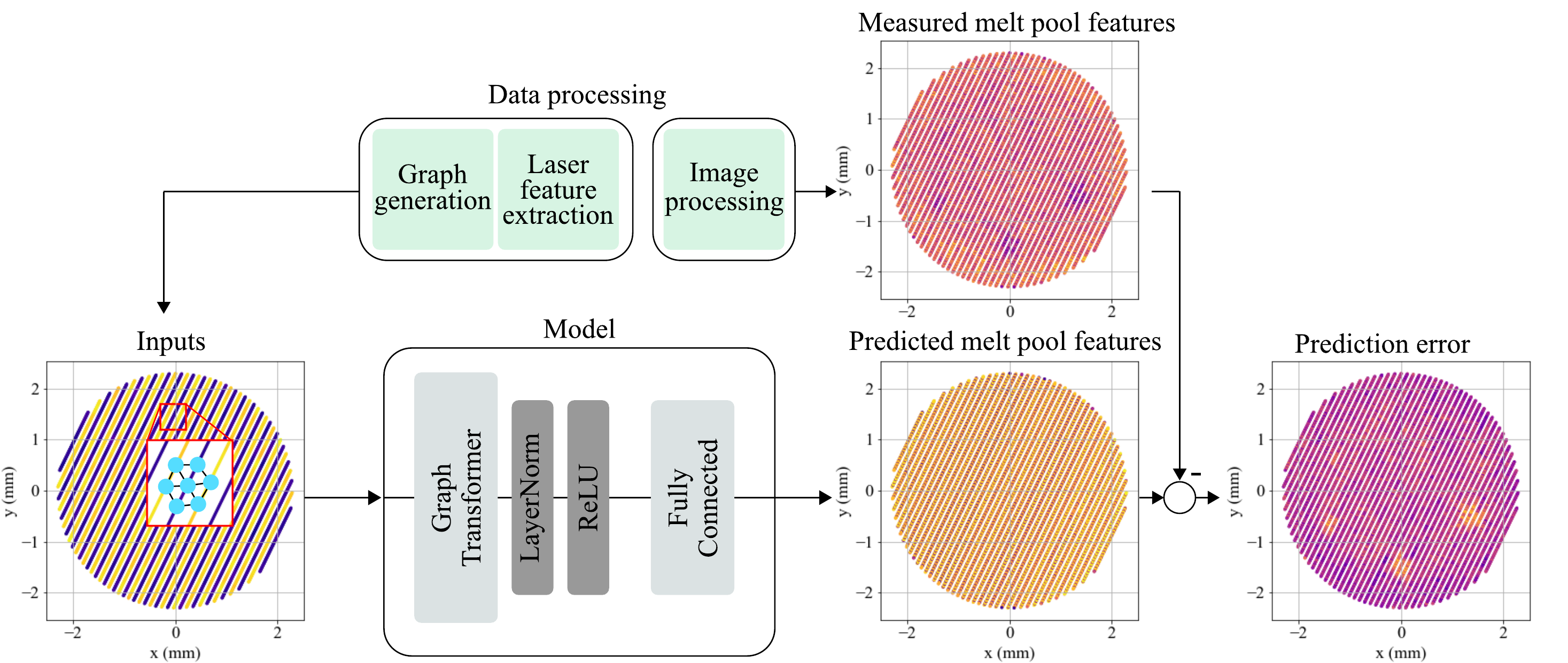}
\end{center}
\caption{The framework for anomaly detection developed in this study.}
\label{fig:GNN}
\end{figure}

The framework for anomaly detection can be seen in Fig. \ref{fig:GNN}. An anomalous observation is defined as a data point that differs considerably from the other data within a set of observations. For instance reduced intensity or excessive spatter shown in Fig. \ref{fig:data}. Therefore, the goal is to learn a score function that can be compared to new observations. Since there are contextual events, such as the laser features and part geometry, this known physics should be incorporated into the model. Hence, we propose to predict the expected sensor observation ($\hat{Y}$) given a set of laser inputs ($X$), $\hat{Y} = f_\theta(X, A)$, where $f_\theta$ is a function mapping of inputs to outputs parametrised by $\theta$ and $A$ is the adjacency matrix. At test time the predicted features are compared to the measured features. This is computed with an error metric given by

\begin{equation}
Z_i = \sum_{j=1}^{n} (\hat{Y}_{i,j} - Y_{i,j}),
\label{eq:1}
\end{equation}

where $Z$ will approach a Gaussian distribution for nominal conditions compared to anomalies. Laser features including scan direction, power, node number and track number were used as model inputs (node features).

As the laser data is correlated both temporally and spatially, a graph was constructed based on the nearest neighbours of each node. These are expected to correlate since a defect can span multiple laser tracks. The input information is fed to a Graph Transformer network \citep{shi2020masked}. This enables the use of edge attributes as well as node features to be represented. In this case, we label edges based on whether they are connected to an adjacent scan track or a not, i.e. temporally or spatially correlated. This is followed by a Fully Connected (FC) post-message passing layer before predicting the expected sensor signature.

The model was trained on nominal processing conditions to minimise the mean squared error (MSE) using the Adam optimiser \citep{kingma2014adam}. Each graph corresponded to a layer of deposited material, with training data on eight graphs totalling 182,450 nodes. The model was evaluated on four graphs known to contain anomalies, where 1,735 nodes were defective and 70,495 were nominal with a data imbalance of 40.6 to 1.

After the error score was computed it was smoothed to remove noise and to enable information propagation in the local neighbourhood of each node. This was performed with a symmetric normalisation and aggregation given by

\begin{equation}
Z' = D^{-\frac{1}{2}} A D^{-\frac{1}{2}} Z,
\label{eq:2}
\end{equation}

where $D$ is the degree matrix.

\section{Results and Discussion}

\subsection{Node Level Predictions}

\begin{figure}[httb]
\begin{center}
\includegraphics[width=9.5cm]{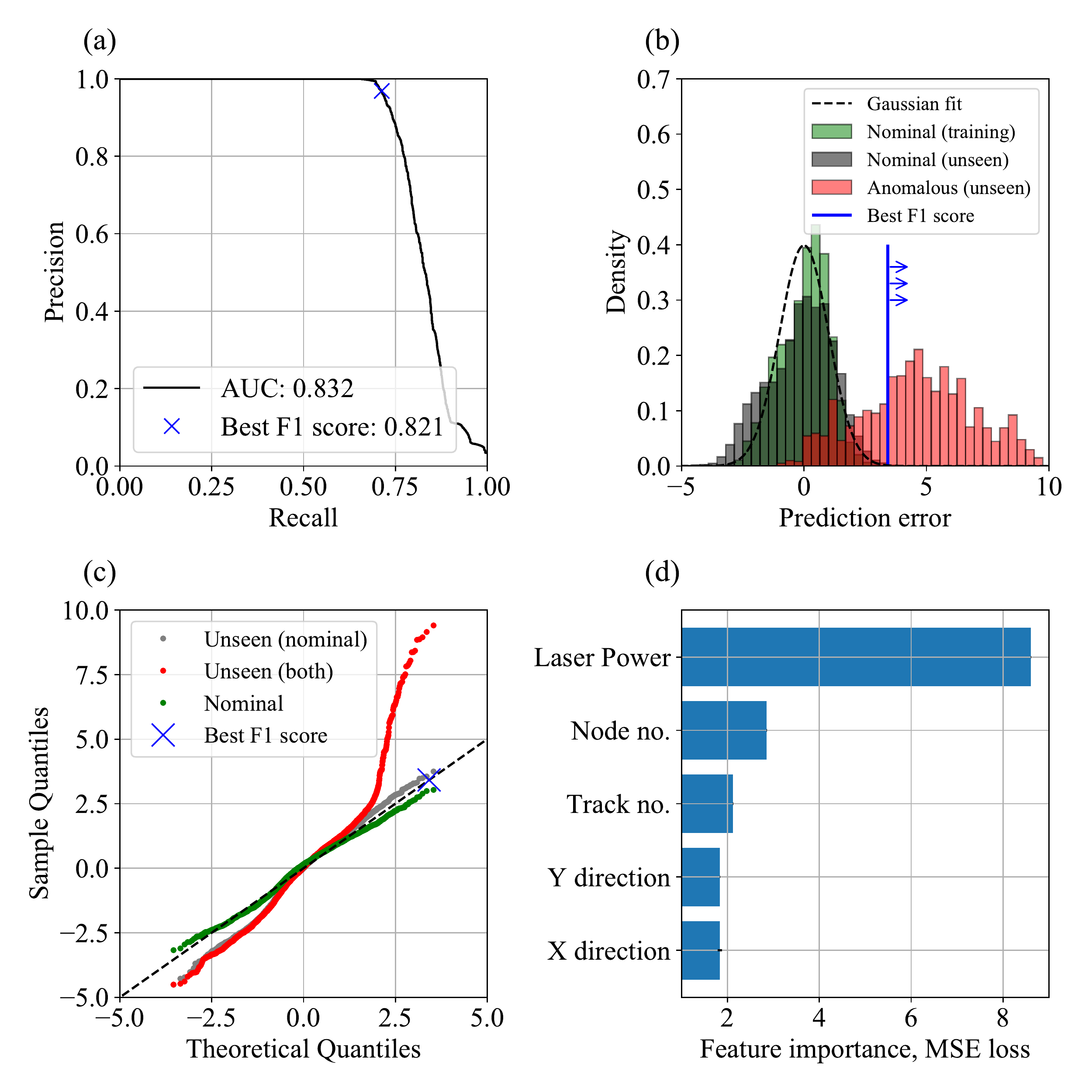}
\end{center}
\caption{(a) Precision-recall curve. (b) Histogram of error for nominal and anomalous data. (c) QQ plot of both nominal and anomalous datasets corresponding to (b). (d) Feature importance of inputs.}
\label{fig:results_pr_curve}
\end{figure}

A threshold was obtained for the outlier model using the precision-recall curve which was chosen due to the data imbalance (Fig. \ref{fig:results_pr_curve}a). The area under the curve, or average precision (AP) was 0.832, while a maximum F1 score was measured as 0.821. Fig. \ref{fig:results_pr_curve}b shows a histogram of nominal and anomalous prediction errors along with the selected threshold from the precision-recall curve.

The prediction error was expected to converge to a Gaussian distribution if the sources of variation are accounted for. This motivates studying the selected threshold from a quantiles perspective (Fig. \ref{fig:results_pr_curve}c). The prediction errors were mapped to a standard Gaussian distribution along with the threshold. This enabled an estimate of how many false positives (FP) might be expected for a given nominal layer. In this case, a threshold of 3.42 resulted in a false positive rate (FPR) of   
around 6 in 10,000 (Table \ref{tab:addlabel}). When compared on the quantile plot of a Gaussian distribution it is expected to be closer to 3 in 10,000. Therefore, an improvement may still be possible though this is in good agreement. However, the quantiles of the unseen data are different, particularly in the negative region. This is likely due to the few graphs used in the dataset (only 8 in training). We believe this can be further improved by including more examples thereby improving the prediction model.

The input feature importance was also evaluated (Fig. \ref{fig:results_pr_curve}d). The laser power was the most important in making predictions. This is due to the laser turning on and off which correlates with the signal in the camera. When the laser turns off, the signal reduces. This information is essential to include since anomalous events, such as a reduced signal, can also be associated with defect formation. The node and track features were less important but provide some contextual information as the laser traverses the  layer heating the component over time. The scanning directions were close to redundant in the predictions.

\subsection{Model Comparisons}

The model was compared to: Graph Attention Networks (GAT) \citep{velivckovic2017graph}, a Graph Convolutional Network (GCN) \citep{kipf2016semi}, a Graph Isomorphism Network (GIN) \citep{xu2018powerful}, and a FC network (Table \ref{tab:addlabel}). Furthermore, an autoencoder (AE) anomaly detection method was included and trained to reconstruct both laser and image features.

The Graph Transformer (Graph-T) model had the highest average precision (AP) and F1 score. The closest performing model (GAT) had similar AUROC but lower AP. We consider the AP a better evaluation in this application due to the data imbalance and add larger weight to the lower FP achieved by the Graph-T model. When comparing the FC with the GNN models (GAT, GCN, GIN), the loss was higher for the GNNs. It is thought that these models might over-smooth the inputs causing some information to be lost in the forward pass, while the FC does not utilise the graph structure until test time. The Graph Transformer can, however, take advantage of both pieces of information.

\begin{table}[htbp]
  \caption{Comparative performance of different models including Graph Transformer with an absolute anomaly metric (Graph-T-A) and without output smoothing (Graph-T-Z).}
   \label{tab:addlabel}%
   \begin{center}
    \begin{tabular}{lrrrrrrrr}
    \multicolumn{1}{c}{\textbf{Model}} & \multicolumn{1}{c}{\textbf{AP}} & \multicolumn{1}{c}{\textbf{AUROC}} & \multicolumn{1}{c}{\textbf{F1}} & \multicolumn{1}{c}{\textbf{FP}} & \multicolumn{1}{c}{\textbf{FN}} & \multicolumn{1}{c}{\textbf{TP}} & \multicolumn{1}{c}{\textbf{TN}} & \multicolumn{1}{c}{\textbf{Loss}} \\
    \midrule
    AE    & 0.808 & 0.958 & 0.800 & 52    & 545   & 1190  & 70443 & 0.0181 \\
    FC    & 0.817 & 0.961 & 0.813 & 65    & 501   & 1234  & 70430 & 0.0025 \\
    GAT   & 0.826 & \textbf{0.963} & 0.815 & 78    & \textbf{489} & \textbf{1246} & 70417 & 0.0147 \\
    GCN   & 0.818 & 0.959 & 0.798 & 95    & 521   & 1214  & 70400 & 0.0273 \\
    GIN   & 0.813 & 0.960 & 0.806 & 71    & 517   & 1218  & 70424 & 0.0148 \\
    Graph-T-A & 0.725 & 0.932 & 0.695 & 386   & 605   & 1130  & 70109 & 0.0024 \\
    Graph-T-Z & 0.246 & 0.833 & 0.338 & 1759  & 1025  & 710   & 68736 & 0.0024 \\
    Graph-T & \textbf{0.832} & \textbf{0.963} & \textbf{0.821} & \textbf{40} & 499   & 1236  & \textbf{70455} & \textbf{0.0024} \\
    \end{tabular}%
    \end{center}
\end{table}%

An absolute error model (Graph-T-A) and an unsmoothed output (Graph-T-Z) were also considered, changing Equation \ref{eq:1} and removing Equation \ref{eq:2} respectively. The absolute error metric had larger error rates since there will be FPs from both sides of the distribution. This confirms that the error signal direction is important to consider in this task.

Removing the smoothing step at the output caused a large decrease in performance at test time. Average precision reduced from 0.832 to 0.246, a decrease of 70\%. This would suggest that smoothing the outputs to minimise noise from predictions is also a critical step to improving the robustness of the anomaly detector.

\section{Conclusion}
\label{gen_inst}

In this study, a GNN was trained to predict the nominal processing conditions from the laser features and positional information represented as a graph. The output prediction was compared to new observations giving a measure of the difference between the two. We found that performing a smoothing step at the output was essential, while the method is robust when compared with a Gaussian distribution. In future work we will test the method on more complex geometries.

\bibliography{iclr2023_conference}
\bibliographystyle{iclr2023_conference}

\end{document}